\documentclass[english, letterpaper, 10pt]{ieeeconf}

\newcommand{\subparagraph}{}

\usepackage[utf8]{inputenc}
\usepackage{amsmath}
\usepackage{amssymb}
\usepackage{graphicx}
\usepackage{cancel}
\usepackage{dsfont}
\usepackage{algpseudocode}
\usepackage{algorithm,algorithmicx}
\usepackage{tikz}
\usepackage{bbm}
\usepackage{dsfont}
\usepackage{graphicx}
\usepackage{dsfont}
\usepackage{graphicx}
\usepackage{subcaption}
\usepackage{multirow}
\usepackage{placeins}

\usepackage{enumerate}
\usepackage{xspace}
\usepackage{romannum}
\usepackage{xcolor,colortbl}
\usepackage{tabularx}
\usepackage[font=scriptsize,labelfont=bf]{caption}

\usepackage{multicol}
\usepackage{wrapfig}
\usepackage{booktabs}

\global\long\def\norma#1{\left\Vert #1\right\Vert }%

\newcommand{\eb}{\mathbf{e}}%

\newcommand{\define}{\triangleq}%
\newcommand{\argThis}[3]
{
	\underset{#2}{#1}
	\left \lbrace #3 \right \rbrace
}

\global\long\def\=#1{\underset{#1}{=}}%

\renewcommand{\tt}{\boldsymbol{\theta}}%
\newcommand{\E}{\mathbb{E}}%
\newcommand{\pr}{\mathbb{P}}%

\global\long\def\diag#1{\text{diag}\left(#1\right)}%

\global\long\def\lim#1{\underset{#1}{\text{lim}}}%

\global\long\def\xx{\mathcal{X}}%

\global\long\def\cc{\mathcal{C}}%

\newcommand{\setout}[1][k]{\Omega^{out}_{#1}}
\newcommand{\setin}[1][k]{\Omega^{in}_{#1}}

\renewcommand{\thesection}{\Roman{section}} 
\renewcommand{\thesubsection}{\thesection.\Roman{subsection}}

\newcommand{\xdomain}[1][k]{\Omega_{\xx_{#1}}}

\newcommand{\Nin}{N_{in}}

\newcommand{\prob}[1]{\ensuremath{\mathbb{P}({#1})}}
\newcommand{\no}{N}
\newcommand{\etab}{\underline{\eta}}

\title{Hybrid Belief Pruning with Guarantees for 
		Viewpoint-Dependent Semantic SLAM}

\author{Tuvy Lemberg and Vadim Indelman %
\thanks{
The authors are with the Department of Aerospace 
Engineering, Technion - Israel Institute of Technology, Haifa 
32000, Israel. tuvy@campus.technion.ac.il, 
vadim.indelman@technion.ac.il. This work was partially supported by US NSF/US- Israel BSF, and by the Israeli Smart Transportation Research Center (ISTRC).
} 
}

\IEEEoverridecommandlockouts
\begin{document}

	\maketitle
	
\begin{abstract}

Semantic simultaneous localization and mapping is a subject of 
increasing interest in robotics and AI that directly influences 
the autonomous vehicles industry, the army industries, 
and more. 
One of the challenges in this field is to obtain 
object classification jointly with robot 
trajectory estimation. 
Considering view-dependent semantic measurements, 
there is a coupling between different classes, 
resulting in a combinatorial number of hypotheses.
A common solution is 
to prune hypotheses that have a sufficiently low probability
and to retain only a limited number of hypotheses.
However, after pruning and renormalization,
the updated probability is overconfident 
with respect to the original probability.
This is especially problematic 
for systems that require high accuracy.
If the prior probability of the classes is independent, the original normalization factor can be computed efficiently without pruning hypotheses.
To the best of our knowledge, this is the first work to present these results.
If the prior 
probability of the classes is dependent, we propose a lower bound on the 
normalization factor that ensures cautious results.
The bound is calculated incrementally and with similar efficiency as in the independent case. 
After pruning and updating based on the bound, this belief is shown empirically to be close to the original belief.

\end{abstract}

	\section{Introduction}

In robotics, the problem of simultaneous localization and mapping (SLAM) 
concerned with estimating robot poses and, concurrently, 
reconstructing the environment observed thus far by the robot
\cite{Smith90, Leonard91tra, DurrantWhyte01, Thrun08book}. The SLAM 
problem is essential for the robot to be able to navigate autonomously in uncertain or unknown environments.  

Recent advances in object recognition and classification 
enabled to 
incorporate object classification and semantic features 
within the SLAM framework,
in order to improve both localization estimations and classification 
\cite{Doherty19icra, Bowman17icra, Tchuiev19iros}. 
In particular, semantic features are considered to be more 
robust and indicative than geometric features 
and can be used in ambiguous and repetitive environments, see 
e.g.~\cite{Feldman18icra}. 
In addition, semantic SLAM provides a high level understanding that is essential for autonomous decision making and advanced robotic capabilities.

Many semantic SLAM works make use of only the maximum likelihood of a 
classifier output \cite{Pillai15rss, Yang19tro, Moreno13cvpr}. This approach reduces the classifier 
output to one discrete variable and cannot represent the uncertainty of the 
classifier given a specific image input. More sophisticated approaches consider 
that classes are not fixed to the maximum likelihood. Yet, many of these approaches assume that 
object's class and object's poses are independent, and as such, solve two separable 
problems (e.g.~\cite{Doherty19icra, Bowman17icra}). However, this corresponds to not
taking into account the inherent 
coupling between classifier output and the viewpoint from which the object is 
seen.

In recent years, several works considered this  coupling 
by using a viewpoint-dependent semantic observation model. In particular, it has been 
shown  that utilizing such a model enhances disambiguation and localization in 
challenging scenarios 
\cite{Tchuiev19iros}.
The resulting belief is hybrid, combining continuous variables 
such as robot and object poses with discrete variables such 
as object classes. 
Moreover, the statistical dependency between the classifier 
output and the viewpoint creates a statistical dependency between classes of 
different objects. Another source for this coupling is a dependent prior probability, i.e.~when the prior over classification variables of all objects is \emph{not} equal to the product of priors on each of the objects. Such a case is more general as it allows to incorporate readily available statistical knowledge (e.g.~high chances to find a keyboard next to a monitor). Therefore, the number of hypotheses is the number of class combinations, which increases \emph{exponentially} with the number of objects. With a few dozen objects and a few hundred classes, the computational complexity becomes prohibitively expensive.

A common way to handle this prohibitive computational 
complexity is by pruning hypotheses that are considered with
the lowest probability. 
Yet, after pruning and renormalization, the resulting belief 
is overconfident in its hypotheses 
with respect to the original belief. 
Overconfident probabilities may lead to reckless and dangerous behavior. 
An autonomous vehicle that considers to continue driving 
according to the overconfident classification may cause fatal 
errors.
Additionally, there is no indication of the significance of pruned 
hypotheses after they have been pruned. Yet, such an indication can be of prime importance for safe autonomy and the decision making process; furthermore,  it may 
be used to trigger a change in maintained hypotheses (e.g.~resurrection of a previously pruned hypothesis).

In this paper, we suggest two methods for maintaining 
probabilities after pruning in a more realistic and conservative 
manner. We show that the in case of an a-priori independent 
class probability, the original normalization factor 
without pruning can be calculated very efficiently.
By obtaining the original normalization factor, we can retrieve the \emph{exact} probability of each hypothesis separately.
If the a-priori probability is dependent, we propose a 
lower bound on the probability of the retained hypotheses.
Applying both methods, provides an indication of the probability 
of the pruned hypotheses. 
Using the first method, the probability of the entire 
pruned set can be computed. Using the second method, 
the probability of the entire pruned set can be bounded from above.
In terms of computational complexity, both methods are similar to pruning.
	\section{Related Work} \label{sec: related work}


It appears that most semantic SLAM methods can be divided into two types. 
In the first type, the semantic observations are considered 
class-dependent and viewpoint-independent. In the second type, 
the semantic observations are considered to be class 
independent and viewpoint dependent.
There are very few studies which consider the semantic 
observations to be both class-dependent and 
viewpoint-dependent. Part of the reason for this is the high 
computational complexity involved in using this model.


Omidshafiei et al. 
\cite{Omidshafiei16arxiv}
proposed a 
Dirichlet distribution as the noisy semantic observation model
for the classifier output.
They developed a Bayesian filter and 
showed that the resulting classification is more robust to ambiguity. 
The semantic observations are assumed to be viewpoint-independent.
Nicholson et al.
\cite{Nicholson18ral}
used two-dimensional object detection and represented objects as three-dimensional
oriented quadratic surfaces. 
The semantic observations are the object shape detections which are considered to be viewpoint-dependent and class-independent; 
Classification took place externally. 
Yang and Scherer, 
\cite{Yang19tro}
proposed object SLAM 
fed by a single image 3D cuboid object detector - the semantic model, which is viewpoint-dependent and class-independent.
In Doherty et al. 
\cite{Doherty19icra}
for each new 
measurement, the marginal posterior of the data association 
was computed. The posterior of the data association was then 
used to compute the posterior of the remaining parameters. 
As a result, they did not have to directly maintain a hybrid belief. 
Moreover, they implemented this method within a factor graph 
in \cite{Doherty20icra}.
In both works, the semantic model is assumed to be 
view independent, thereby avoiding 
the mentioned high computational complexity.

Both viewpoint-dependent and class-dependent models were recently studied.
A neural network was used by Kopitkov and Indelman to learn 
a viewpoint-dependent measurement model of CNN classifier 
output features \cite{Kopitkov18iros}. 
That study focused on the learning procedure of both viewpoint-dependent and class-dependent semantic model, and how to incorporate it as a factor in a factor graph framework.
Feldman and Indelman 
\cite{Feldman18icra}
learned a viewpoint-dependent model of  semantic measurements, 
and represent the model as a Gaussian process.
Tchuiev and Indelman  
\cite{Tchuiev18ral}
developed a method for sequential reasoning about the posterior 
uncertainty of a semantic model. 
This method was later applied to autonomous planning scenarios \cite{Tchuiev21arxiv}.
Feldman and Indelman 
\cite{Feldman20rss_ws} replaced categorical class variables with 
latent continuous object representations.
This representation provides the potential to represent semantic 
information in a more meaningful way than only the object's class.

Bowman et al. 
\cite{Bowman17icra}
showed that a joint hybrid belief of trajectory, 
data association, landmark locations, and landmark class can be 
maximized through expectation-maximization (EM).
In the case of scenarios containing many classes and objects, 
the number of hypotheses through which the expectation must 
pass is too large and therefore is still infeasible.  
Tchuiev et al.
\cite{Tchuiev19iros} 
used a hybrid belief to infer also data association. 
Data association hypotheses with small probabilities were pruned. 
They used a semantic model within the factor graph, 
improving both classification and localization.
However, there is no guarantee that the updated belief after 
pruning will be close to the original belief.
Furthermore, due to the high computational complexity of using such a model, they were limited to scenarios with very few classes and objects.
	\section{Problem Formulation and Notations\label{Sec:problem formulation}}
Suppose a robot travels in an unknown environment and receives observations from various objects 
as it moves.
Let $N$ denote the number of objects the robot observes and
$M$  the number of possible classes.
Denote the robot's pose at time-step $ k $ by $ x_k $,
and its trajectory from the beginning to time-step $ k $
by $x_{1:k}=\left\lbrace x_1,\ldots, x_k \right\rbrace$.
Let $x^o_n$ represent the pose of the $n$th object, and $X^o = \left\lbrace x^o_1, \ldots ,x^o_N \right\rbrace$ the poses of all $N$ objects.
The class of the $ n $th object is $ c_n $, 
and the concatenation of all objects' classes is
$C = \left\lbrace c_1, \dots, c_N \right\rbrace$.

The robot receives observations
$z_k = \left\lbrace z_{k,1},\ldots, z_{k,\no_k} \right\rbrace$
at the $k$th time step,
where $ \no_k $ is the number of objects observed 
at time-step $ k $.
In general, it is unknown which observation $z_{k,j}$ 
corresponds to which of the objects. 
Let the data association (DA) random variable $\beta_{k,j} = n$ 
represent the event that observation $z_{k,j}$  
corresponds to the 
$n$th object, and, 
$\beta_k =  \{\beta_{k,1},\ldots, \beta_{k,\no_k}\}$
is the concatenation of all DAs at time-step $k$.
In this paper, we assume that DA is known 
and will extend to unknown DA in the future.
Each observation $z_{k,j}$ consists of a geometric part 
$z_{k,j}^g $ and a semantic part $z_{k,j}^s$, that are assumed to be independent on each other. The observations model is given by
\begin{multline}	
	\pr\left(
		z_{k,j} \mid x^o_{\beta_{k,j}},x_i,c_{\beta_{k,j}}
	\right)
	= \\
	\pr\left(
		z^s_{k,j} \mid x^o_{\beta_{k,j}},x_k,c_{\beta_{k,j}}
	\right)
	\pr\left(
		z^g_{k,j} \mid x^o_{\beta_{k,j}},x_k
	\right),
	\label{eq:observation model}
\end{multline}
where 
$
\pr\left(
z^s_{k,j} \mid x^o_{\beta_{k,j}},x_k,c_{\beta_{k,j}}
\right)  
$ is the viewpoint-dependent semantic model, and
$ 
\pr\left(
z^g_{k,j} \mid x^o_{\beta_{k,j}},x_k
\right)
$
is the geometric model, and both are assumed to be given.
Additionally, the actions $ a_{0:k-1} $  and the motion model $\pr\left(x_k 
\mid x_{k-1}, a_{k-1}\right)$ are also assumed to be given. 

Define $\xx_k = \lbrace X^o, x_{1:k} \rbrace$  
as the concatenation of all unknown continuous random variables;  we shall refer to it as the state.  Define  history at time-step $k$ as $H_k = \lbrace z_{1:k}, a_{1:k-1} \rbrace$. The joint hybrid belief over $\xx_k$ and $C$ is defined as follows
\begin{equation}
	b\left[\xx_k,C\right] \define
	\pr\left(\xx_k,C\mid H_k\right).
	\label{eq:belief definition}
\end{equation}
We present a recursive derivation of the belief. 
Based on Bayes' theorem, the last observation can be pulled from the history
\begin{equation}
	\pr\left(\xx_k,C\mid H_k\right) \!\! = \!
	\eta_k
	\pr\!\left(z_k\mid\xx_k,C,H_k^-\right)\!
	\pr\!\left(\xx_k,C\mid H_k^-\right)\!,
	\label{eq:belief_bayes}
\end{equation}
 where 
$
\eta_k\define \pr\left(z_k\mid H_k^-\right)^{-1}
$ 
 is the normalization factor and 
 $H_k^-$ is the the history without the last observations, $z_k$.
The observations are dependent only on the state at the current 
time, therefore
\begin{equation}
	\pr\left(z_k\mid\xx_k,C,H_k^-\right) 
	 =
 	\prod_{j=1}^{\no_k}
	 \pr\left(
	 z_{k,j}\mid x_k,c_{\beta_{k,j}},  x^o_{\beta_{k,j}}
	 \right)	 
\end{equation}
Applying chain rule, the last term in \eqref{eq:belief_bayes} can be written as
\begin{equation}
	\pr\left(\xx_k,C\mid H_k^-\right)  
	=
	\pr\left(x_k\mid a_{k-1},x_{k-1}\right)
	\pr\left(\xx_{k-1},C\mid H_{k-1}\right).
\end{equation}
This process can be repeated recursively, resulting in the following formulation
\begin{multline}
	\tilde{b}_k\left[\xx_k,C\right]  
	=
	\pr_0\left(C\right)
	\pr_0\left(\xx_0\right)
	\prod_{i=1}^k
	\pr\left(x_i\mid a_{i-1},x_{i-1}\right) \\
	\times
	\prod_{j=1}^{\no_k}
	\pr\left(
	z_{i,j}\mid x_i,c_{\beta_{i,j}},  x^o_{\beta_{i,j}}
	\right)
	,
	\label{eq:belief propagation total}
\end{multline}
where $ c_{\beta_{i,j}} $ is the element of $ C $
corresponding to index $ \beta_{i,j} $.
By normalizing \eqref{eq:belief propagation total}
we get the belief 
which is the joint probability of $ \xx_k $ and $ C $
\begin{equation}
	b_k\left[\xx_k,C\right] 
	=
	\eta_{1:k}\tilde{b}_k\left[\xx_k,C\right]
	,
\end{equation}
where the normalization factor  $ \eta_{1:k} $ is given by 
\begin{equation}
	\eta_{1:k}^{-1} =
	\sum_{C\in\cc}
	\intop_{\xdomain}
	\tilde{b}_k\left[\xx_k, C\right]
	d\xx_k.
	\label{eq:normalization factor definition}
\end{equation}
The belief can be rearranged so that all observations 
related to the same object are grouped together.
Consider $ I\left(n,k\right) $ 
as the set of indices 
$\left<i,j\right>$
of observations associated with the $ n $th object
that were observed up to time-step $ k $.
$ I\left(n,k\right) $ can be defined mathematically as follows 
\begin{equation}
	I\left(n,k\right)
	=
	\left \lbrace 
		\left<i,j\right>
		: \beta_{i,j} = n, 1 \leq i \leq k, \: 1 \leq j \leq \no_i
	\right \rbrace.
\end{equation}
The belief can then be reordered as follows
\begin{multline}
	b\left[\xx_k,C\right] 
	=
	\eta_{1:k}
	\pr_0\left(C\right)
	\pr_0\left(X^0\right)
	\left(
		\prod_{i=1}^k
		\pr\left(
			x_{i} \mid  a_{i-1},x_{i-1}
		\right)
	\right)
	\\
	\times
	\prod_{n=1}^N
	\prod_{i,j \in I\left(n,k\right)}
	\pr\left(
		z_{i,j}\mid x_{n}^o,x_i,c_{n}
	\right).
	\label{eq:belief separated by object}
\end{multline}

Using a viewpoint-dependent semantic model, it is evident that classes of different objects are dependent. 
There are two reasons for this. First, the prior probability of the 
classes  may be dependent, i.e., $\pr_0\left(C\right) \neq \prod_{n=1}^N \pr_0\left(c_n\right)$, for example, near crosswalks we expect to see pedestrians. 
Second, because of the viewpoint-dependent semantic model, the classes are dependent on the poses, which are in turn dependent on each other, so the classes are dependent.

In general, to infer $ \xx_k $ and $ C $, 
it is necessary to go through all possible combinations of classes,
even when using an efficient 
algorithm such as expectation maximization.
The number of combinations, or hypotheses, is $ M^N $. 
As a result, all algorithms that consider all of the hypotheses
run at least at $ O \left(M ^ N\right) $. 
Because of this, pruning the vast majority of hypotheses 
is essential.
Using the naive pruning approach, after the pruning and 
renormalization, it is impossible to determine 
whether the pruned hypotheses maintain a high probability. 
In naive pruning, the probabilities of the remaining 
hypotheses are assumed to sum to one, 
meaning that the probabilities of the pruned 
hypotheses are assumed to be zero.

Yet, it is irresponsible to assume that the probabilities of the pruned set are zero.
Consider that a robot operates in a human environment, and 
the robot can do some potentially dangerous things only if it is 
confident that no human is nearby. By using naive pruning, 
the robot will be overconfident in the retained hypotheses. 
In the event that the correct hypothesis was pruned, because there is no indication that the correct hypothesis was pruned, the robot may assume that there is no human nearby and proceed to take the dangerous action.
In contrast, if the robot obtains the exact probability that the correct hypothesis is pruned, or at least a lower bound on this probability, the robot will know that it lacks sufficient confidence to take the dangerous action and that the correct hypothesis was pruned.

\begin{figure}[H]
	\centering
	\includegraphics[width=.25\textwidth]
	{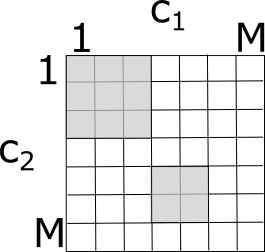}
	\caption
	{Illustration of the pruned set $ \setout $ in white and the  retained set $ \setin $, in gray in a simple case of two objects. A hypothesis consists of a class for each object. It is considered that $ |\setin| \ll |\setout|$.
	}
	\label{fig:setin setout illustration}
\end{figure}
	 \section{Approach \label{sec:Approach}}

We propose two alternative approaches to evaluating the normalization factor after pruning.
We consider two cases. 
In the first case, the prior probability of the classes is independent, i.e.
$ \pr_0\left(C\right) = \prod_{n=1}^N \pr_0\left(c_n\right)$. 
Our study will demonstrate that in this case the original normalization factor can be calculated with the same computational efficiency as the pruning version of the normalization factor.
Because we have the original normalization factor, we can query the original probability of each hypothesis. In this case, we know the real probabilities of the remaining hypotheses, and if they are low, we should replace them.

In the second case, the prior probability of the classes is dependent, i.e. $ \pr_0\left(C\right) \neq \prod_{n=1}^N \pr_0\left(c_n\right)$.
In this case, we propose an upper bound on the probabilities of the pruned hypotheses. Using this bound, we derive a lower bound on the normalization factor.
Using an upper bound on the probability of the pruned hypotheses, we can determine whether there is a high probability hypothesis within the pruned set.
This bound is also computed with the same computational efficiency.

In both cases, we denote the set of maintained hypotheses by $ \setin $  and the rest of the (pruned) hypotheses  by $ \setout $ (see illustration in Fig.~\ref{fig:setin setout illustration}).


	
\subsection{Independent Class Prior}
In order to simplify notation, the unnormalized belief  
\eqref{eq:belief propagation total} 
can be divided into two parts, 
one containing the semantic observations, 
and the other containing the rest.
\begin{multline}
	\tilde{b}_k \left[ \xx_k, C \right]
	=
	\\
	\left(
		\pr_0 \left( X^0 \right)
		\prod_{i=1}^k
		\pr\left(x_i \mid a_{i-1}, x_{i-1}\right)
		\prod_{j=1}^{\no_i}
		\pr\left(
			z^g_{i,j} \mid x^o_{\beta_{i,j}},x_i
		\right)
	\right) 
	\\
	\times
	\left(
		\pr_0 \left( C\right)
		\prod_{i=1}^k \prod_{j=1}^{\no_i}
		\pr\left(
			z^s_{i,j} \mid x^o_{\beta_{i,j}},x_i,c_{\beta_{i,j}}
		\right)
	\right)
	.
	\label{eq: belief separated to semantic and not}
\end{multline}
Define the geometric unnormalized belief as follows
\begin{equation}
	\tilde{b}^g_k \left[\xx_k\right] 
	\define 
	\pr_0\left(\xx_0\right)
	\prod_{i=1}^k
	\pr\left(x_i\mid a_{i-1},x_{i-1}\right)
	\prod_{j=1}^{\no_i}
	\pr\left(
	z_{i,j}^g \mid x_{\beta_{i,j}}^o, x_i
	\right),
	\label{eq:geometric belief}
\end{equation}
where $ H_k^g $ is the history without the semantic observations.
The unnormalized belief is equal to the probability of 
$ \xx_k \mid H_k^g$ 
multiplied by the normalization factor $ \eta^g_{1:k}$.
By substituting \eqref{eq:geometric belief}
into 
\eqref{eq: belief separated to semantic and not}, 
and reordering the equation as in 
\eqref{eq:belief separated by object},
one obtains
\begin{equation}
	\tilde{b}_k\left[\xx_k,C\right] 
	=
	\tilde{b}^g_k \left[\xx_k\right]
	\pr_0\left(C\right)
	\prod_{n=1}^N
	\prod_{i,j \in I\left(n,k\right)}
	\pr\left(
	z^s_{i,j}\mid x_{n}^o,x_i,c_{n}
	\right).
	\label{eq:belief geometric part separated}
\end{equation}
Further simplifying the belief, 
we define $ \psi_k$ as follows
\begin{equation}
	\psi_k(n,c,\xx_k) 
	\define
	\prod_{i,j\in I\left(n,k\right)}
	\pr\left(z^s_{i,j} \mid x_i, x_n^o, c_n = c\right)
	.
	\label{eq: psi}
\end{equation}
It is the likelihood of the object's class $ c_n=c $ and the state,
given the semantic observations 
associated with the $ n $th object.
Substituting $ \psi_k $ into 
\eqref{eq:belief geometric part separated}
results in 
\begin{equation}
	\tilde{b}_k\left[\xx_k,C\right] 
	=
	\tilde{b}^g_k \left[\xx_k\right]
	\pr_0\left(C\right)
	\prod_{n=1}^N
	\psi_k(n,c_n,\xx_k)
	.
	\label{eq:simplified belief}
\end{equation}
The normalization factor is given in 
\eqref{eq:normalization factor definition},
and by substituting \eqref{eq:simplified belief},
we obtain
\begin{equation}
	\eta_{1:k}^{-1}=
	\sum_{C\in\cc} 
	\intop_{\xdomain} 
	\tilde{b}^g_k \left[\xx_k\right]
	\pr_0\left(C\right)
	\prod_{n=1}^N
	\psi_k(n,c_n,\xx_k)
	d\xx_k.
	\label{eq:normalization factor implicit}
\end{equation}
The above integration, \eqref{eq:normalization factor implicit},
can be represented as expectation over $ \xx_k \mid H_k^g $ up to the scale of the normalization factor
$ \eta^g_{1:k}$ 
of the geometric belief,
results in the following
\begin{equation}
	\eta_{1:k}^{-1}= 	
	(\eta^g_{1:k})^{-1}
	\sum_{C\in \cc}\pr_0\left(C\right)
	\E_{\xx_k \mid H_k^g}
	\left[
	\prod_{n=1}^N	
	\psi_k(n,c_n,\xx_k)
	\right]
	.
	\label{eq:normalization expectation}
\end{equation}
Since the normalized belief is required only for the probabilities of classes, the state will be marginalized later. Thus, we could apply the same expectation as in \eqref{eq:normalization expectation}, and the normalization factor 
$ \tilde{b}^g_k $ will be eliminated.
Until now, the derivation has been general and does not 
assume independent priors and therefore applies also to the 
dependent case.

By using the independent property of the prior of $ C $,
and reorganizing 
\eqref{eq:normalization factor implicit},
we obtain
\begin{equation}
	\eta_{1:k}^{-1}= 
	\E_{\xx_k \mid H_k^g}
	\left[
		\sum_{c_1=1}^M \dots \sum_{c_N=1}^M	
		\prod_{n=1}^N
		\pr_0\left(c_n\right)
		\psi_k(n,c_n,\xx_k)
	\right]
	.
	\label{eq:normalization factor 2}
\end{equation}
Our \emph{key observation} is that it is possible to substitute the sum and the product, rewriting  \eqref{eq:normalization factor 2}  as
\begin{equation}
	\eta_{1:k}^{-1}= 
	\E_{\xx_k \mid H_k^g}
	\left[	
	\prod_{n=1}^N
	\pr_0\left(c_n=c\right)
	\sum_{c=1}^M
	\psi_k(n,c,\xx_k)
	\right]
	,
	\label{eq:normalization factor final independent}
\end{equation}
since each $ c_n $ is independent of the other, inside the expectation.
Using \eqref{eq:normalization factor final independent},
the computing complexity of the normalization factor is 
significantly reduced. 

Let us compare the running time of the naive approach in 
\eqref{eq:normalization expectation}
and the efficient approach in 
\eqref{eq:normalization factor final independent}. 
In both cases, integration of the state and calculation of 
$ \psi_{k}(n,c,\xx_k)$
is required. Since this calculation is common, it is not included in the analysis to follow.  
Consider that the expectation over $\xx_k$  is approximated by $ N_s $ samples drawn from 
$ b_k^g[\xx_k]$.
Using the naive approach, one must sum the unnormalized probabilities for all hypotheses, as shown in 
\eqref{eq:normalization expectation}.
Therefore, the naive approach runs in $ O(M^N\cdot N_s) $. In contrast, 
the computational complexity of the efficient approach \eqref{eq:normalization factor final independent} is 
$ O\left(N\cdot M\cdot N_s\right) $.

To summarize, using the exact normalization factor \eqref{eq:normalization expectation}, the \emph{exact} probability can be computed for each of the maintained hypotheses in $ \setin $.  
Moreover, the exact probabilities of hypotheses in $ \setin $, provide the exact probability of $ \prob{C \in \setout} $, which can indicate if the correct hypothesis was pruned.  
In Fig. \ref{fig:setin setout illustration}, and illustration of $ \setin $ and $ \setout $
is provided.

	\subsection{Dependent Class Prior}

\begin{figure*}
	\vspace*{-1cm}
\begin{subfigure}[b]{0.23\linewidth}
	\includegraphics[width=\linewidth]
		{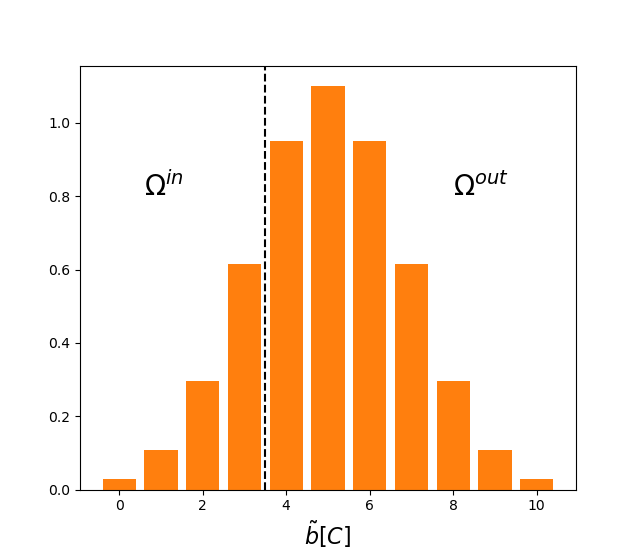}		\caption{\label{fig:pf:bound_illustration:a}
			Unnormalized belief
		}
\end{subfigure}
\begin{subfigure}[b]{0.23\linewidth}
	\includegraphics[width=\linewidth]
	{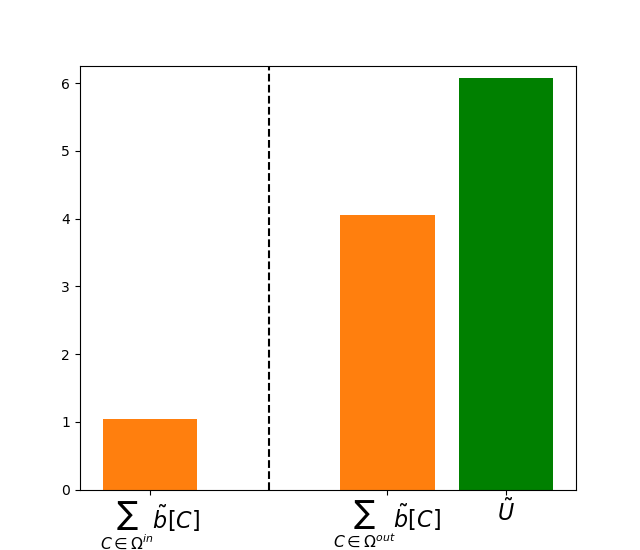}		\caption{\label{fig:pf:bound_illustration:b}
	Unnormalized sums and bound}
\end{subfigure}
\begin{subfigure}[b]{0.23\linewidth}
	\includegraphics[width=\linewidth]
	{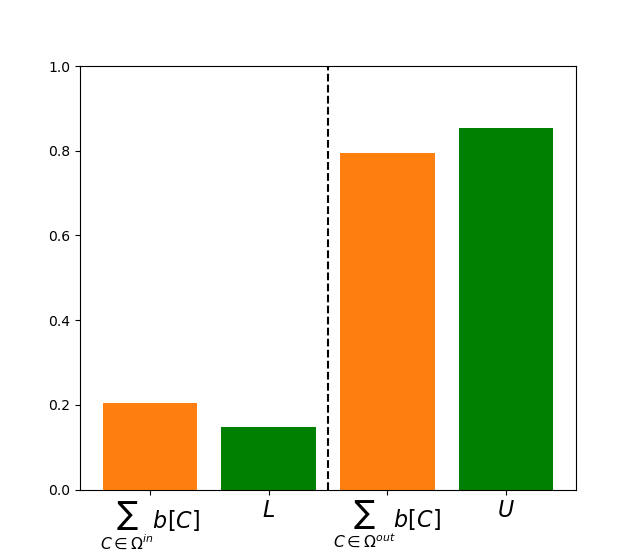}		\caption{\label{fig:pf:bound_illustration:c}
	Normalized sums and bound}
\end{subfigure}
\begin{subfigure}[b]{0.23\linewidth}
	\includegraphics[width=\linewidth]
	{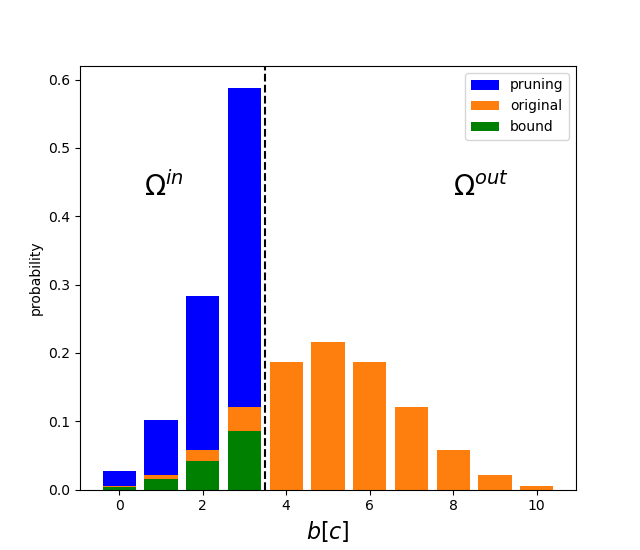}		\caption{\label{fig:pf:bound_illustration:d}
	Normalized belief}
\end{subfigure}

\caption
{	
The bound process is illustrated in the following figures.
\eqref{fig:pf:bound_illustration:a}
Unnormalized belief separated to 
$\setin[~]$ in the left (before the dashed line) and 
$\setout[~]$ in the right.
In \eqref{fig:pf:bound_illustration:b},
$\sum_{C\in\setin[~]}\tilde{b}[C]$ 
and $\tilde{U}$ are computed. 
$\sum_{C\in\setout[~]}\tilde{b}[C]$ is not computed.
In \eqref{fig:pf:bound_illustration:c}, the sums bound are 
normalized using $ \etab $.
A comparison is shown \eqref{fig:pf:bound_illustration:d} 
between the original normalized probabilities,
the bound and the naive pruning approach.
Using naive pruning can lead to absurd overconfident 
probabilities. 
Furthermore, it cannot indicate that a better hypothesis is in
the pruned set $\setout[~]$.
In contrast, if there is a high probability that the correct 
hypothesis was pruned, then using the upper bound $ U $ will 
provide an indication of this and will allow for a more 
conservative and realistic normalization of the retained 
hypotheses. 
Suppose a hypothesis in $\setin[~]$ has probability 
$\eta \tilde{b}[C]$ higher than $U$, 
then we are guaranteed that this hypothesis has higher 
probability than any hypothesis in $\setout[~]$.
}
\label{fig: bound illustration}
\end{figure*}

In this case we consider the prior of the classes is dependent. 
Suppose we have an upper bound 
$\tilde{U}_k$ that fulfills the following inequality
\begin{equation}
	\sum_{C\in\setout}
	\tilde{b}_k\left[C\right]  
	\leq 
	\tilde{U}_k. 
	\label{eq:unnormalized bound}
\end{equation}
The normalization factor can be bounded from below as
\begin{equation}
	\eta_{1:k}
	=
	\left(	
	\sum_{C\in\cc} 	
	\tilde{b}_k\left[C\right]  
	\right)^{-1}
	\geq
	\left(
	\sum_{C\in\setin} 	
	\tilde{b}_k\left[C\right]
	+
	\tilde{U}_k
	\right)^{-1}
	\define
	\etab_{1:k}
	,
	\label{eq: normalization factor bound}
\end{equation}
where $ \etab_{1:k}$ is the normalization factor induced by the bound.
The probabilities of $ C $ can be bounded from below
\begin{equation}
	b\left[C\right] = 
	\eta_{1:k}\tilde{b}\left[C\right]
	\geq
	\etab_{1:k}\tilde{b}\left[C\right],
	\label{eq:lower bound on belief}
\end{equation}
A lower bound for the class's probability is a significant result. 
It enables an autonomous robot to take cautious 
actions, and indicates whether $ \prob{C \in\setout} $ 
is high.

Furthermore, we can also bound the probability  
$ \pr\left(C\in\setout\mid H_{k}\right)  $
from above.
\begin{multline}
	\pr\left(C\in\setout\mid H_{k}\right) 
	=
	\sum_{C\in\setout}
	b\left[C\right]
	\\
	=1-	\sum_{C\in\setin}
		b\left[C\right] 
	\leq
	1-	\sum_{C\in\setin}
	\etab_{1:k}\tilde{b}\left[C\right]
	.
	\label{eq:normalized bound 1}
\end{multline}
This bound is also useful, since it indicates if it is likely that 
the correct hypothesis was pruned.
Fig. \ref{fig: bound illustration}  illustrates the derivation of the bound clearly.

Next, we will derive the bound and present a method for 
updating it efficiently.
We begin with the simpler version, where the state is assumed 
to be known, and then we generalize it to the case where it is 
unknown. 

	\subsubsection{Derivation of $\tilde{U}_k$}
We will begin by considering that the $\xx_k$ is known 
and then extend it to an unknown $\xx_k$.
The unnormalized belief of $ C $ 
can be derived from degenerating 
\eqref{eq:belief propagation total}
\begin{equation}
\tilde{b}_k\left[C\right] 
=
\pr_0(C)
\prod_{i=1}^k
\prod_{j=1}^{\no_k}
\pr\left(
z^s_{i,j}\mid x_i,c_{\beta_{i,j}},  x^o_{\beta_{i,j}}
\right).
\label{eq:belif unknown class}
\end{equation}
We obtain the following belief by reorganizing 
\eqref{eq:belif unknown class}
as was done in \eqref{eq:belief separated by object}, 
and plugging in $ \psi_k(n,c,\xx_k)  $
\begin{equation}
	\tilde{b}_k\left[C \right] 
	=
	\pr_0\left(C\right)
	\prod_{n=1}^N
	\psi_k(n,c_n,\xx_k).
	\label{eq: belief class separated}
\end{equation}
According to \eqref{eq:unnormalized bound}, 
we need to bound \eqref{eq: class need to bound} from above
\begin{equation}
	\sum_{C\in\setout}
	\tilde{b}_k\left[C\right] 
	=
	\sum_{C\in\setout}
	\pr_0\left(C\right)
	\prod_{n=1}^N
	\psi_k(n,c,\xx_k) 
	.
	\label{eq: class need to bound}
\end{equation}
Using the Cauchy-Schwarz inequality,
\eqref{eq: class need to bound} 
can be bounded from above as follows
\begin{equation}
	\sum_{C\in\setout}
	\tilde{b}_k\left[C\right]
	\leq
	\left(
	\sum_{C\in\setout}
	\pr_0^2\left(C\right)
	\right)^{1/2}
	\left(
	\sum_{C\in\setout}
	\prod_{n=1}^N
	\psi_k^2(n,c,\xx_k) 
	\right)^{1/2}  
	.
	\label{eq: class bound cauchy schwartz}
\end{equation}
Furthermore, we can use the Hölder's inequality,
which is an extension of the 
Cauchy–Schwarz inequality. 
The Hölder's inequality states that for any two vectors 
$ u, v $ in some inner product space, 
the following inequality holds
\begin{equation}
	\left| \left< u, v\right> \right| 
	\leq 
	\norma{u}_{q_1} \cdot \norma{v}_{q_2},
\end{equation}
for any $ q_1, q_2 \geq 1 $ satisfy
$ \dfrac{1}{q_1} + \dfrac{1}{q_2} = 1 $,
where $ \norma{\cdot}_q $ is the $ L(q) $ norm. 
Therefore, 
\begin{equation}
	\tilde{U}_k \left(\xx_k\right)
	=
	\left(
	\sum_{C\in\setout}
	\pr_0^{q_1}\left(C\right)
	\right)^{1/q_1}
	\left(
	\sum_{C\in\setout}
	\prod_{n=1}^N
	\psi_k^{q_2}(n,c,\xx_k) 
	\right)^{1/q_2}  
	.
\end{equation}
	\subsubsection{Efficient update of $\tilde{U}_k$ 
	\label{subsection: bound update}}

Two events can cause an update: a new observation is 
received, and a change in the composition of the $\setin, 
\setout$.
In the former case, the bound should be updated incrementally 
like the belief.
In the latter case, we avoid starting from scratch and re-use 
previous calculations.

Suppose that the number of elements in $ \setin $
is limited to 
$\Nin = \left| \setin\right| \ll \left|\setout\right|$.
A sum over $\setout$ can be replaced with 
sum over $\cc$ minus sum over $\setin$.
Using this attribute, the bound can be rewritten as follows 
\begin{multline}
	\tilde{U}_k \left(\xx_k\right)
	=
	\left(
		\sum_{C\in\cc}
		\pr_0^{q_1}\left(C\right)
		-
		\sum_{C\in \setin}
		\pr_0^{q_1}\left(C\right)
	\right)^{1/q_1} 
	\\
	\times
	\left(
		\sum_{C\in\cc}
		\prod_{n=1}^N
		\psi_k^{q_2}(n,c,\xx_k) 
		-
		\sum_{C\in\setin}
		\prod_{n=1}^N
		\psi_k^{q_2}(n,c,\xx_k) 
	\right)^{1/q_2}.
	\label{eq:cl:unnormalized bound}
\end{multline}
In order to simplify $ \tilde{U}_k $ , the sums inside 
\eqref{eq:cl:unnormalized bound} can be defined as following 
\begin{align}
	S^0_\cc 
	& \define
	\sum_{C\in\cc}
	\pr_0^{q_1}\left(C\right)
	\label{eq: sum 0 on C}
	\\
	S^0_{\setin} 
	& \define
	 \sum_{C\in\setin}
	\pr_0^{q_1}\left(C\right)
	\label{eq: sum 0 in}
	\\
	S^{\psi_k}_{\setin} \left(\xx_k\right)
	& \define
	\sum_{C\in\setin}
	\prod_{n=1}^N
	\psi_k^{q_2}(n,c_n, \xx_k)
	\label{eq: sum psi in}
	\\	
	S^{\psi_k}_\cc \left(\xx_k\right)
	& \define 
	\sum_{C\in\cc}
	\prod_{n=1}^N
	\psi_k^{q_2}(n,c, \xx_k)	.
	\label{eq: sum psi on C}
\end{align}
In \eqref{eq: sum psi in}, the class of the $ n $th object 
is the $ n $th element of the vector $ C $.
In light of definitions 
(\ref{eq: sum 0 on C}-\ref{eq: sum psi on C}),
the bound \eqref{eq:cl:unnormalized bound} can be rewritten as follows
\begin{equation}
	\tilde{U}_k \left(\xx_k\right)
	= 
	\left( S_\cc^0    - S_{\setin}^0\right)^{1/q_1}
	\left(
		S_\cc^{\psi_k}\left(\xx_k\right) -
		S_{\setin}^{\psi_k}\left(\xx_k\right)
	\right)^{1/q_2}
	.
	\label{eq:class bound simplified}
\end{equation}

\subsubsection*{Update due to new observation}

Observing new observation $ z_{k} $ will change
$ S^{\psi_k}_\cc $, 
$ S^{\psi_k}_{\setin} $ and 
$ \psi_k $.
As before, we exclude the time complexity of computing $ \psi_k $.
The dependency on $ \xx_k $ 
is omitted here for simplicity of notation,
keeping in mind that $ \psi_k $ is dependents on $ \xx_k $.
The computation time for the update of
$ \psi_k$ is 
$ O(M\cdot \no_k) $.
For the update of $ S_{\setin}^{\psi_k} $, 
define
\begin{equation}
	\varphi_k(C) 
	\define 
	\prod_{n=1}^N
	\psi_k(n,c=c_n,\xx_k)
	\:\:\: C \in \setin.
	\label{eq:varphi}
\end{equation}
It is clear that 
\begin{equation}
	S_{\setin}^{\psi_k} = \sum_{C\in \setin} \varphi_k(C). 
\end{equation}
$ \varphi_k(C) $ updates as follows
\begin{multline}
	\varphi_k(C) =
	\varphi_{k-1}(C)
	\prod_{j=1}^{\no_k}
	\pr^{q_2}\left(
		z^s_{k,j} \mid
		x_{\beta_{k,j}}^o,x_{k},c=c_{\beta_{k,j}}
	\right)
	.
\end{multline}
The computational time for updating $  \varphi_k(C) $ 
is $ O(\no_k) $, it should be done for all $ C\in\setin $,
therefore, the computation time for updating
$ S_{\setin}^{\psi_k} $
is
$ O(\no_k \Nin) $.

The explicit formulation of $S^{\psi_k}_\cc$ is given by
\begin{multline}
	S^{\psi_k}_\cc  
	= 
	\sum_{C\in\cc}
	\prod_{n=1}^N
	\psi_k(n,c,\xx_k)
	\\	
	=
	\sum_{c_1=1}^M \cdots \sum_{c_N=1}^M
	\prod_{n=1}^N
	\psi_k(n,c=c_n,\xx_k).
	\label{eq:class S psi}
\end{multline}
The sum and the product can be substituted for each other in this case
\begin{equation}
	S^{\psi_k}_\cc = 
	\prod_{n=1}^N
	\sum_{c_n=1}^M
	\psi_k(n,c=c_n,\xx_k).
	\label{cl:eq:S tutal}
\end{equation} 
Define 
\begin{equation}
s^{\psi_k}_n 
\define 
\sum_{c=1}^M
\psi_k(n,c,\xx_k)
.
\label{eq: s_n psi k}
\end{equation} 
\eqref{eq: s_n psi k} can be inserted to 
$ S^{\psi_k}_\cc $
\begin{equation}
	S_\cc^{\psi_k} = \prod_{n=1}^N s^{\psi_k}_n.
	\label{eq:class psi efficient}
\end{equation}
The update affects only $ s^{\psi_k}_n$ for $n \in \beta_k $. The computation time for updating 
$ s^{\psi_k}_n$ is 
$ O(M) $ and for 
$ S_\cc^{\psi_k} $ is 
$ O\left( M \cdot \no_k \right) $.
The final computation time for an observation-based update is 
$ O(\no_k\max(M, \Nin)) $.

\subsubsection*{Update due to change in $ \setin, \setout $}
The second type of update is caused by a change in the composition of $ \setin, \setout $, and 
affects only $ S_{\setin}^0,  S_{\setin}^{\psi_k}$.
Consider adding a new hypothesis, $ C^a $,
and removing existing hypothesis $ C^r $ 
from $ \setin $. 
The update will be performed as follows
\begin{align}
	S_{\setin}^0 
	& \leftarrow 
	S_{\setin}^0 + \pr_0^{q_1}(C^a) 
	\\
	\varphi(C^a)
	& = 
	\prod_{n=1}^N
	\psi_k(n,c=c_n^a) 
	\\ 	
	S_{\setin}^{\psi_k} 
	& \leftarrow	
	S_{\setin}^{\psi_k} + \varphi(C^a).
\end{align}
The update involves computing $ \varphi(C^a)$, 
which is $ O(N) $.
Now the removal of $ C^r $ is done by
\begin{align}
	S_{\setin}^0 
	& \leftarrow 
	S_{\setin}^0 - \pr_0^{q_1}(C^r) \\
	S_{\setin}^{\psi_k} 
	& \leftarrow	
	S_{\setin}^{\psi_k} - \varphi(C^r),
\end{align}
which is done in $ O(1) $.
	\subsection{Belief over unknown classes and state	\label{sec:unknown-state}}
The unnormalized belief over the classes and the state can be taken from 
\eqref{eq:simplified belief}.
According to \eqref{eq:unnormalized bound}, 
$ \tilde{U}_k $
should satisfy
\begin{equation}
	\tilde{U}_k 
	\geq
	\sum_{C\in\setout} 
	\intop_{\xdomain} 
	\tilde{b}^g_k \left[\xx_k\right]
	p_0\left(C\right)
	\prod_{n=1}^N
	\psi_k(n,c_n,\xx_k)
	d\xx_k.
	\label{eq:inequality bound state case} 
\end{equation}
Integration can be replaced by expectation 
over $ \xx_k \mid H_k^g$, 
as done in \eqref{eq:normalization factor 2},
\begin{equation}
	\tilde{U}_k 
	\geq
	\E_{\xx_k \mid H_k^g} \left[
	\sum_{ C \in \setout}
	p_0 \left(C\right)
	\prod_{i=1}^k
	\prod_{j=1}^{n_i}
	\psi_k(n,c_n,\xx_k)
	\right].
	\label{st:eq:to be bound}
\end{equation}
The term inside the expectation of 
\eqref{st:eq:to be bound}
is the unnormalized belief of $ C\mid \xx_k $
given in \eqref{eq: belief class separated}.
Therefore, it can also be bounded using the bound in 
\eqref{eq:class bound simplified}, 
thus
\begin{equation}
	\tilde{U}_k = 
	\E_{\xx_k \mid H_k^g} \left[
	\left( S_\cc^0    - S_{\setin}^0\right)^{1/q_1}
	\left(
		S_\cc^{\psi_k} \left(\xx_k\right)
		-
		S_{\setin}^{\psi_k} \left(\xx_k\right)
	\right)^{1/q_2}
	\right].
	\label{eq:st: bound}
\end{equation}
$ \left( S_\cc^0    - S_{\setin}^0\right)^{1/q_1} $
is not dependent on $ \xx_k $ 
so it can be taken out of the expectation.
The time complexity of computing the bound is the same as before multiplied by the number of samples $ N_s $.
	\section{Experiments \label{sec: experiment}}

In this section, we evaluate our methods using synthetic simulations.
It is important to note that we do not propose 
a new viewpoint-dependent model, 
but only suggest that given such a model is utilized, 
our method can reduce computational complexity substantially

A 2D environment is assumed with a robot that moves around, 
observing the objects that it encounters along the way.
Each object has orientation and class type. 
Objects are randomly positioned throughout the environment, 
and their classes are randomly selected.
Initially, the robot does not know where the objects are placed or to which class they belong.
The geometric observations are bearing measurements 
and the semantic observations are synthetic classifier output.
Based on the geometric measurements and the motion model, 
the geometric belief is evaluated.
The geometric belief is assumed to be Gaussian.
To calculate the full SLAM graph, 
we used the GTSAM library with Python wrapper, 
resulting in a mean and a covariance of the geometric belief. 
Throughout all the simulations, 
100 samples are drawn from the geometric belief.

For the robot's prior pose, process noise, and geometric observation noise, the covariances were
$\Sigma_0= \diag{0.01, 0.01, 0.001} $, 
$\Sigma_p= \diag{0.3, 0.3, 0.03} $, 
and 
$\Sigma_m= \diag{0.03} $, respectively.
The dependent prior probability of $ C $ is randomly selected using the following procedure. A matrix $ A $ of size
$ \underset{N}{\underbrace{M\times M\dots\times M}} $ was allocated.
The elements of $ A $ was sampled from the uniform distribution $ U[0.001, 1] $. Then $ A $ was divided by the sum of its elements.
Thus, the prior probability of $ C $ is given by
$ \pr_0(C) = A[C] $.
In the case of an independent prior, this procedure was used for each marginal.

\begin{figure}[H]
	\centering
	\includegraphics[width=0.5\linewidth]
	{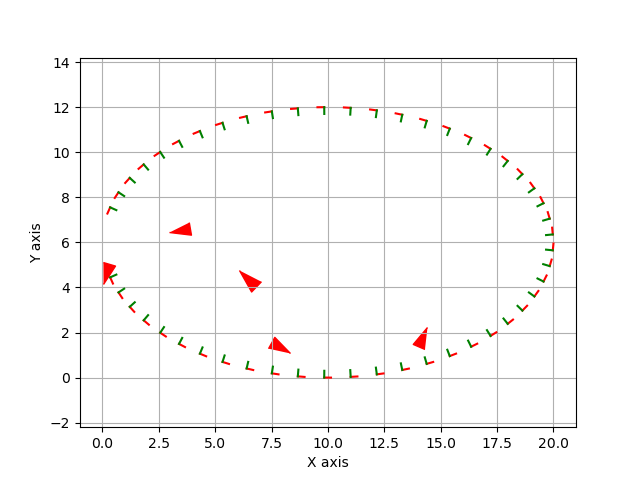}
	\caption
	{
Visualization of the scene. 
The triangles in red represent objects that are oriented. 
The angles and classes were chosen at random. 
The robot moves counterclockwise, starting from the lower left side. 
	}
	\label{fig:fig:sceneclass scene}
\end{figure}

In Fig.~\ref{fig:fig:sceneclass scene} we can see the scheme of
the scene.

The distribution of the semantic observations is logit-normal distribution,
thus,
\begin{equation}
	z_{k,j} \sim  
	LN\left(\mu\left(x^o_{\beta_{k,j}}, x_k,c \right), \Sigma \right) ,
	\label{eq:semantic observation model}
\end{equation}
where $ \Sigma=I_M\sigma_s^2 $ and $ \sigma^s=0.015 $.
The same distribution is used in equation 
\cite{Tchuiev21arxiv} as well.
It is possible to interpret this distribution as the output of a 
classifier since the samples are vectors of positive elements that 
sum to 1.
For simplicity of notation, consider that $ z_{k,j} $ is associated with the $ n $th object, thus $ \beta_{k,j}=n $.
The semantic measurement model is defined as follows
 \begin{equation}
 	\mu\left(x^o_n, x_k,c\right) = 
 	\begin{cases}
 		e_c  
 		\cdot 
 		h\left(x^o_n, x_k\right)
 		& c\in[1,M-1]
 		\\
 		0_{M-1}& c=M
 	\end{cases},
 	\label{eq:semantic mean}
 \end{equation}
\begin{equation}
	h\left(x^o_n, x_k\right) = 
	\left(1-\cos\left(\theta\right)\right)
	\min\left\{ 
	\frac{1}{dist(x_k, x_n^o)}
	,\frac{1}{2}
	\right\},
	\label{eq:semantic coeff}
\end{equation}
where $ \eb_c\in \mathbb{R}^{M-1} $is a vector of zeros with a single element at the $ c $th position equal to one, 
and $ dist(\cdot, \cdot) $ is the euclidean distance.
$ \theta $ is the relative angle, calculated from the relative pose $ x^{rel} = x_k\ominus x_n^o $.
The mean 
$ \mu\left(x^o_n, x_k,c\right) $
is characterized by the following features:
under the the hypothesis $ c=M $, the mean of $ z_{k,j} $
is $ \frac{1}{M} $. For the other hypotheses,
as the camera gets closer to the object
and becomes oriented towards the front of the object, the probability that the right element of $ z_{k,n} $ will respond to  object's class is increased.

We compare between the original probability without pruning, 
the naive pruning approach, the bound approach, and pruning using the original normalization in the independent prior case.

The sets $ \setin, \setout $ are predefined and are constant. 
The size of $ \setin $ is $\Nin = 8 $.
A true hypothesis 
$ C^{(t)} =  (c_1^{(t)},\ldots,c_N^{(t)})$ 
is predefined as well. 
The semantic observations are sampled from the probability
$ \pr\left(z_{k,n} \mid x_n^o,x_k,c_n^{(t)}\right) $
described in 
\eqref{eq:semantic observation model},
\eqref{eq:semantic mean},
and
\eqref{eq:semantic coeff}.

The results are divided into two cases: 
when  $ C^{(t)} \in \setin $
and where $ C^{(t)} \in \setout $.
In order to determine whether the system is confident in its best hypothesis, we look at the maximum probability in $ \setin $,
$ \argThis{\text{max}}{C \in \setin }{b[C]} $, over time.
The most likely hypothesis is the same across all methods because the difference between them is only the normalization factors, but different probability is provided by each approach.
The same sets $ \setin,  \setout $ used for all methods.

\begin{figure}
	\begin{subfigure}[b]{0.45\linewidth}
		\includegraphics[width=\linewidth]
		{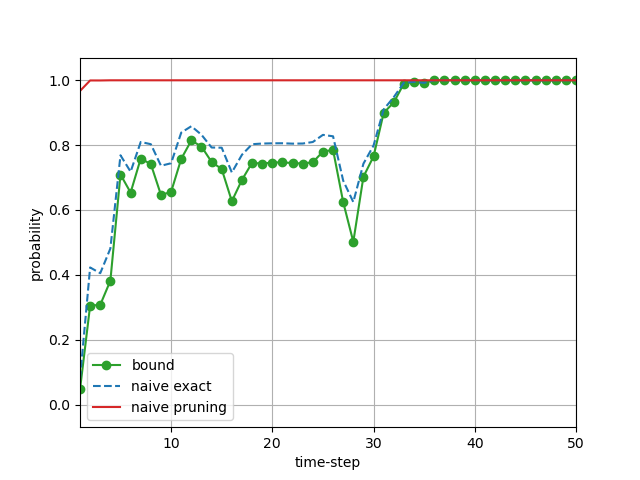}
		\caption{\label{fig: dependent C in}}
	\end{subfigure}
	\begin{subfigure}[b]{0.45\linewidth}
		\includegraphics[width=\linewidth]
		{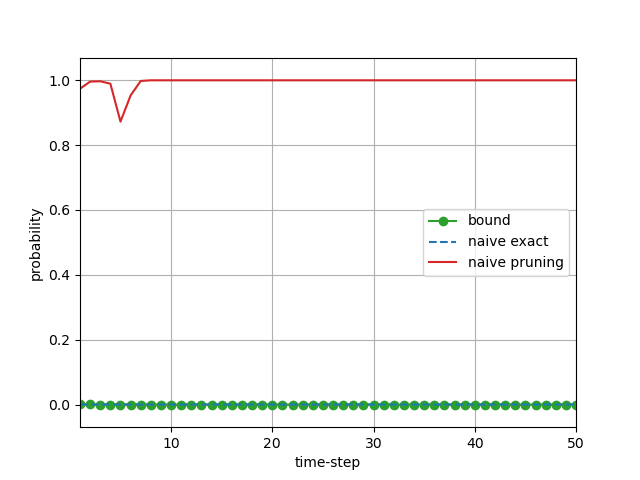}
		\caption{\label{fig: dependent C out}}
	\end{subfigure}
		\caption{The maximal probability of the maintained hypotheses. Considering an \emph{dependent} prior, where 
		\textbf{(a)} 
		$ C^{(t)} \in \setin$, and \textbf{(b)} 
		$ C^{(t)} \in \setout$.
	}
\end{figure}

Both Fig. \ref{fig: dependent C in} and Fig. \ref{fig: dependent C out}
illustrate the scenario of classes with dependent priors.
The naive exact belief, naive pruning, and bound are compared.
The number of objects is 5 and number of classes is 3, the total number of hypotheses is $ 3^5 = 243 $. Even when considering such a small number of classes and objects, the number of hypotheses is not so small.

Fig. \ref{fig: dependent C in} shows the case for 
$ C^{(t)} \in \setin$.  The naive pruning approach is very confident in 
its hypothesis from the beginning to the end. The naive exact approach, which provides the exact probability of this hypothesis, is less confident. 
The bound, which bounds the exact probability from below 
\eqref{eq:lower bound on belief}, following the exact approach throughout the simulation. 
Although the bound is a lower bound on the exact belief, it remains close to the exact belief even when the latter reaches a high probability.

Fig. \ref{fig: dependent C out}, shows the case for $ C^{(t)} \in \setout$.
Throughout the simulation, the exact probability is close to zero and therefore also the bound. 
By contrast, the naive pruning approach is too confident, assuming that the correct hypothesis is  in $ \setin $.

\begin{figure}
	\begin{subfigure}[b]{0.45\linewidth}
			\includegraphics[width=\linewidth]
		{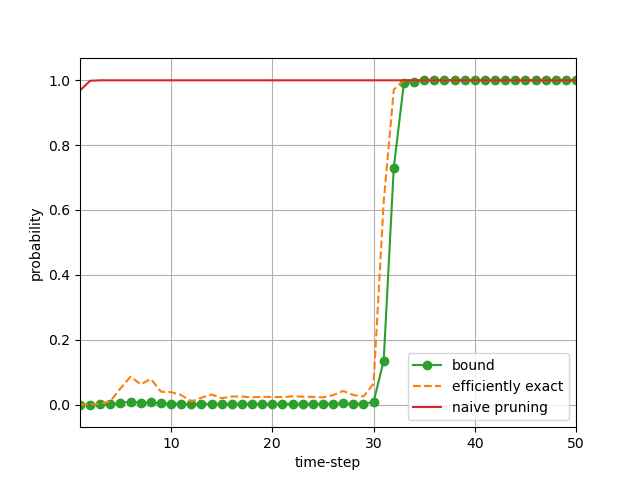}
		\caption{\label{fig:independent C in }}
	\end{subfigure}
\begin{subfigure}[b]{0.45\linewidth}
	\includegraphics[width=\linewidth]
	{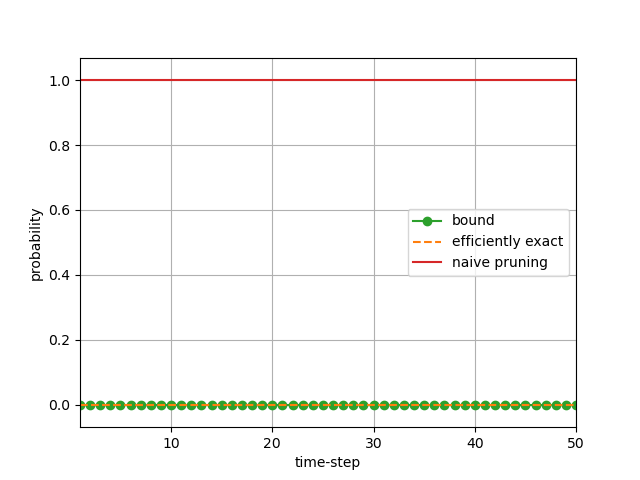}
	\caption{\label{fig: independent C out}}
\end{subfigure}
	\caption{The maximal probability of the maintained hypotheses. Considering an \emph{independent} prior, where 
		\textbf{(a)} 
		$ C^{(t)} \in \setin$, and \textbf{(b)} 
		$ C^{(t)} \in \setout$. 
	}
\end{figure}

The independent prior case is considered in figures
Fig. \ref{fig:independent C in } and 
Fig. \ref{fig: independent C out}.
For the independent prior case,  we used $ N=5 $ and $ M=100 $.
In this scenario the number of hypotheses is $ 10^{10} $.
In this case, we compare the efficient exact belief with naive pruning and the bound approach. Despite the fact that the bound is not required in this case, we wish to evaluate how it will work when many classes are involved.

In this case, the results appear to be the same as in the dependent case.
In Fig. \ref{fig:independent C in }, the naive approach is overconfident. 
In the exact belief we see that only after several time-steps the belief gain confidence,
and the bound is follows the exact belief.
According to  Fig. \ref{fig: independent C out}, the naive approach is absurdly 
overconfident, while the exact belief and the bound approach indicate that 
the probability is zero.

\begin{figure}
	\begin{subfigure}[b]{0.45\linewidth}
		\includegraphics[width=\linewidth]
		{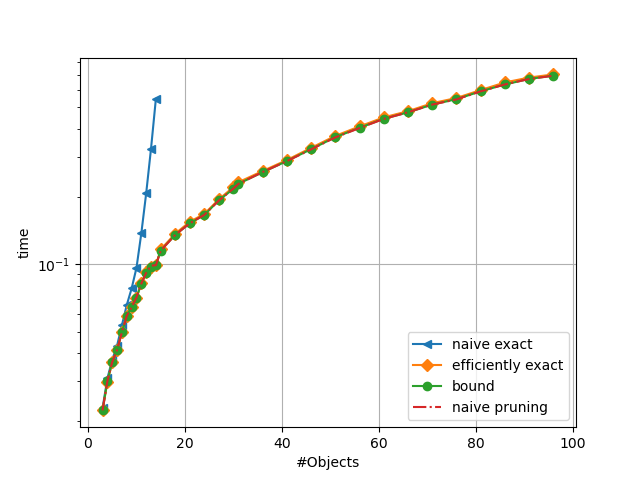}
		\caption{\label{fig: running time vs N}}
	\end{subfigure}
	\begin{subfigure}[b]{0.45\linewidth}
		\includegraphics[width=\linewidth]
		{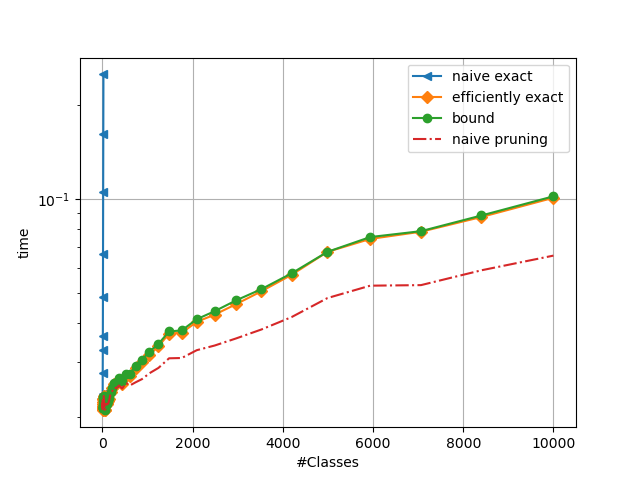}
		\caption{\label{fig: running time vs M}}
	\end{subfigure}
	\caption{\textbf{(a)} 
		Running time versus number of objects $ N $. The number of classes is $ M=2 $. The computational complexity of both the bound and the efficient method for the independent case original  are the same as pruning. \textbf{(b)} Running time versus number of classes $ M $. The number of object is $ N=3 $. Here, the naive pruning running time is slightly better. 
	}
\end{figure}

In figures Fig. \ref{fig: running time vs N} and 
Fig. \ref{fig: running time vs M}
we compare the runtime of 
naive pruning, naive exact, the efficient exact and the bound approach. 
For each simulation, one hundred trails were performed and the mean value of the runtime was taken.

In Fig. \ref{fig: running time vs N}, show the runtime versus the number of objects $ N $. The number of classes is $ M=2 $. 
It appears that the efficient exact and the bound approach have the same runtime as the naive pruning approach. 
In Fig.  \ref{fig: running time vs M},
the runtime is plotted against the number of classes $ M $.
The number of objects is $ N=3 $.
In comparison to a naive pruning approach, there is a slight increase in runtime for the efficient exact and bound approaches.
Given the significant improvement in accuracy and reliability, we consider the efficient exact and bound approaches to be worthwhile.

	\section{Conclusions \label{sec: conclutions}}
In the context of semantic SLAM, we explore  the pruning procedure often used when the size of the alphabet of the hybrid belief is exponentially large,
in the case where the semantic observation 
model is dependent on the relative position of the object.
In this scenario, the number of hypotheses is exponential
in the number of objects, and therefore, 
not feasible in a real-time framework.
Due to this reason, the pruning procedure is used,
but following pruning and renormalization, 
the resulting belief is overconfident, 
disregards the possibility 
that the true hypothesis is pruned. 
When the prior over classification variables of different objects is independent, we demonstrated that the normalization factor can be calculated efficiently, allowing us to query the exact probability of each hypothesis individually. 
In the dependent case, using our method, we bound the probability that the true hypothesis was pruned, and, as a result, we obtain a more accurate and conservative estimate of the original belief, 
with the guarantee that the original belief over classification variables without pruning is equal or higher than our belief after pruning.
Future research might explore the minimization of the upper bound $U_k$ with 
respect to $\setin$ as a method to decide which 
hypotheses to maintain and which to prune.

	\bibliographystyle{IEEEtran}
	\bibliography{main.bib}

\begin{thebibliography}{10}
\providecommand{\url}[1]{#1}
\csname url@samestyle\endcsname
\providecommand{\newblock}{\relax}
\providecommand{\bibinfo}[2]{#2}
\providecommand{\BIBentrySTDinterwordspacing}{\spaceskip=0pt\relax}
\providecommand{\BIBentryALTinterwordstretchfactor}{4}
\providecommand{\BIBentryALTinterwordspacing}{\spaceskip=\fontdimen2\font plus
\BIBentryALTinterwordstretchfactor\fontdimen3\font minus
  \fontdimen4\font\relax}
\providecommand{\BIBforeignlanguage}[2]{{%
\expandafter\ifx\csname l@#1\endcsname\relax
\typeout{** WARNING: IEEEtran.bst: No hyphenation pattern has been}%
\typeout{** loaded for the language `#1'. Using the pattern for}%
\typeout{** the default language instead.}%
\else
\language=\csname l@#1\endcsname
\fi
#2}}
\providecommand{\BIBdecl}{\relax}
\BIBdecl

\bibitem{Smith90}
R.~Smith, M.~Self, and P.~Cheeseman, ``Estimating uncertain spatial
  relationships in {R}obotics,'' in \emph{Autonomous Robot Vehicles}, I.~Cox
  and G.~Wilfong, Eds.\hskip 1em plus 0.5em minus 0.4em\relax Springer-Verlag,
  1990, pp. 167--193.

\bibitem{Leonard91tra}
J.~Leonard and H.~Durrant-Whyte, ``Mobile robot localization by tracking
  geometric beacons,'' \emph{{IEEE} Trans. Robot. Automat.}, vol.~7, no.~3, pp.
  376--382, 1991.

\bibitem{DurrantWhyte01}
H.~Durrant-Whyte, S.~Majunder, S.~Thrun, M.~de~Battista, and S.~Scheding, ``A
  {B}ayesian algorithm for simultaneous localization and map building,'' in
  \emph{Proceedings of the 10th International Symposium of Robotics Research},
  2001.

\bibitem{Thrun08book}
S.~Thrun, \emph{Simultaneous Localization and Mapping}.\hskip 1em plus 0.5em
  minus 0.4em\relax Berlin, Heidelberg: Springer Berlin Heidelberg, 2008, pp.
  13--41.

\bibitem{Doherty19icra}
K.~Doherty, D.~Fourie, and J.~Leonard, ``Multimodal semantic slam with
  probabilistic data association,'' in \emph{2019 international conference on
  robotics and automation (ICRA)}.\hskip 1em plus 0.5em minus 0.4em\relax IEEE,
  2019, pp. 2419--2425.

\bibitem{Bowman17icra}
S.~Bowman, N.~Atanasov, K.~Daniilidis, and G.~Pappas, ``Probabilistic data
  association for semantic slam,'' in \emph{IEEE Intl. Conf. on Robotics and
  Automation (ICRA)}.\hskip 1em plus 0.5em minus 0.4em\relax IEEE, 2017, pp.
  1722--1729.

\bibitem{Tchuiev19iros}
V.~Tchuiev, Y.~Feldman, and V.~Indelman, ``Data association aware semantic
  mapping and localization via a viewpoint-dependent classifier model,'' in
  \emph{IEEE/RSJ Intl. Conf. on Intelligent Robots and Systems (IROS)}, 2019.

\bibitem{Feldman18icra}
Y.~Feldman and V.~Indelman, ``Bayesian viewpoint-dependent robust
  classification under model and localization uncertainty,'' in \emph{IEEE
  Intl. Conf. on Robotics and Automation (ICRA)}, 2018.

\bibitem{Pillai15rss}
S.~Pillai and J.~Leonard, ``Monocular slam supported object recognition,'' in
  \emph{Robotics: Science and Systems (RSS)}, 2015.

\bibitem{Yang19tro}
S.~Yang and S.~Scherer, ``Cubeslam: Monocular 3-d object slam,'' \emph{IEEE
  Transactions on Robotics}, vol.~35, no.~4, pp. 925--938, 2019.

\bibitem{Moreno13cvpr}
R.~F. Salas-Moreno, R.~A. Newcombe, H.~Strasdat, P.~H. Kelly, and A.~J.
  Davison, ``Slam++: Simultaneous localisation and mapping at the level of
  objects,'' in \emph{IEEE Conf. on Computer Vision and Pattern Recognition
  (CVPR)}, 2013, pp. 1352--1359.

\bibitem{Omidshafiei16arxiv}
S.~Omidshafiei, B.~T. Lopez, J.~P. How, and J.~Vian, ``Hierarchical bayesian
  noise inference for robust real-time probabilistic object classification,''
  \emph{arXiv preprint arXiv:1605.01042}, 2016.

\bibitem{Nicholson18ral}
L.~Nicholson, M.~Milford, and N.~S{\"u}nderhauf, ``Quadricslam: Dual quadrics
  from object detections as landmarks in object-oriented slam,'' \emph{{IEEE}
  Robotics and Automation Letters (RA-L)}, vol.~4, no.~1, pp. 1--8, 2018.

\bibitem{Doherty20icra}
K.~J. Doherty, D.~P. Baxter, E.~Schneeweiss, and J.~J. Leonard, ``Probabilistic
  data association via mixture models for robust semantic slam,'' in \emph{2020
  IEEE International Conference on Robotics and Automation (ICRA)}.\hskip 1em
  plus 0.5em minus 0.4em\relax IEEE, 2020, pp. 1098--1104.

\bibitem{Kopitkov18iros}
D.~Kopitkov and V.~Indelman, ``Robot localization through information recovered
  from cnn classificators,'' in \emph{IEEE/RSJ Intl. Conf. on Intelligent
  Robots and Systems (IROS)}.\hskip 1em plus 0.5em minus 0.4em\relax IEEE,
  October 2018.

\bibitem{Tchuiev18ral}
V.~Tchuiev and V.~Indelman, ``Inference over distribution of posterior class
  probabilities for reliable bayesian classification and object-level
  perception,'' \emph{{IEEE} Robotics and Automation Letters (RA-L)}, vol.~3,
  no.~4, pp. 4329--4336, 2018.

\bibitem{Tchuiev21arxiv}
------, ``Epistemic uncertainty aware semantic localization and mapping for
  inference and belief space planning,'' \emph{arXiv preprint
  arXiv:2105.12359}, 2021.

\bibitem{Feldman20rss_ws}
Y.~Feldman and V.~Indelman, ``Towards self-supervised semantic representation
  with a viewpoint-dependent observation model,'' in \emph{Workshop on
  Self-Supervised Robot Learning, in conjunction with Robotics: Science and
  Systems (RSS)}, July 2020.

\end{thebibliography}

\end{document}